\documentclass[letterpaper]{article} 
\usepackage[preprint]{aaai2027}
\usepackage[hyphens]{url}  
\usepackage{graphicx} 
\usepackage{natbib}  
\usepackage{caption} 
\usepackage{algorithm}
\usepackage{algorithmic}

\usepackage{tabularx}
\usepackage{amsmath,amssymb,amsfonts}

\usepackage{multirow}
\usepackage{amsthm}
\usepackage{hyperref}
\usepackage{cleveref}
\usepackage{makecell}

\usepackage{xcolor}
\usepackage{colortbl}
\usepackage{graphicx}

\usepackage{newfloat}
\usepackage{listings}
\DeclareCaptionStyle{ruled}{labelfont=normalfont,labelsep=colon,strut=off}
\floatstyle{ruled}
\newfloat{listing}{tb}{lst}{}
\floatname{listing}{Listing}

\definecolor{BestCell}{RGB}{245,226,250}
\newcommand{\bestcell}[1]{\cellcolor{BestCell}\textbf{#1}}

\definecolor{ieeeblue}{RGB}{0,72,138}
\definecolor{bandgray}{RGB}{240,240,240}
\definecolor{highlight}{RGB}{226,236,248}

\usepackage{booktabs}

\title{Route by Kinematics, Act by Observation: Kinematics-Supervised \\ Expert Routing in MoE-Augmented VLA}

\author {
    Tianhang Yang\textsuperscript{\rm 1,\rm 2}\equalcontrib,
    Yanze Zheng\textsuperscript{\rm 1}\equalcontrib,
    Junjie Wang\textsuperscript{\rm 1},
    Wei-Bin Kou\textsuperscript{\rm 1}\corresponding,
    Ruotong Li\textsuperscript{\rm 1,\rm 2},
    Yujiu Yang\textsuperscript{\rm 1}\corresponding
}

\affiliations {
    \textsuperscript{\rm 1}Tsinghua Shenzhen International Graduate School, Tsinghua University, Shenzhen, China\\
    \textsuperscript{\rm 2}Pengcheng Lab, Shenzhen, China\\
     thu\_tianhangyang@163.com, 221840196abc@gmail.com, wangjunjie@sz.tsinghua.edu.cn, wbkouqvb@sz.tsinghua.edu.cn, lirt\_cs@163.com, yang.yujiu@sz.tsinghua.edu.cn
}

\begin{document}

\maketitle

\begin{abstract}
While MoE augments VLA via expert specialization, router suffers from ineffective expert routing owing to the kinematic heterogeneity of actions across manipulation tasks and, even worse, the unavailability of the kinematic signals at inference time. In this work, we first observe that most semantically distinct manipulation tasks reduce to multiple kinematic archetypes. Motivated by this finding, we propose Kinematics-supervised explicit routing (KinRT), a new paradigm that shifts from implicit, observation-driven expert routing to explicit, kinematics-guided expert dispatching. Specifically, we perform kinematic clustering on action trajectories into multiple kinematically coherent groups, whose IDs serve as ground truth to supervise the training of the router; at inference time, the router dispatches experts only using visual-language observations, without any reliance on action kinematics. KinRT actually introduces an asymmetric bridging mechanism that distills the task kinematics from the action space in training into the observation space at inference. In addition, to assess KinRT's cross-platform generalization, we build an economical, Do-It-Yourself robot (DIYRobot) platform from scratch using 3D-print technology ($<$ 2,000USD). Extensive experiments demonstrate KinRT's superiority over both dense and MoE-featured VLAs by more than 23.26\% on RoboTwin benchmark and 20.27\% on our introduced DIYRobot platform. Our work's homepage is available at: \href{https://gleeacast.github.io/KinRT/}{https://gleeacast.github.io/KinRT/}.
\end{abstract}

\section{Introduction}
\label{sec:intro}

Vision-Language-Action (VLA) models have emerged as a unified architecture for robot manipulation tasks by fusing visual perception, language reasoning, and action generation within a single framework~\cite{team2024octo,kim2024openvla,black2024pi_0,liu2025rdt,zhang2026atomicvla}. Yet, as the task repertoire scales up, the underlying kinematics across tasks become increasingly heterogeneous. Compressing such disparate kinematic patterns into a shared model space inevitably induces gradient interference and poor generalization~\cite{yu2020gradient, liu2021conflict}. Mixture-of-Experts (MoE) \cite{fedus2022switch,ref:expertchoice} offers a promising solution by enabling different experts to specialize in distinct kinematic regimes, thereby alleviating parameter competition while maintaining a unified architecture.

\begin{figure}[t]
\includegraphics[width=1\linewidth]{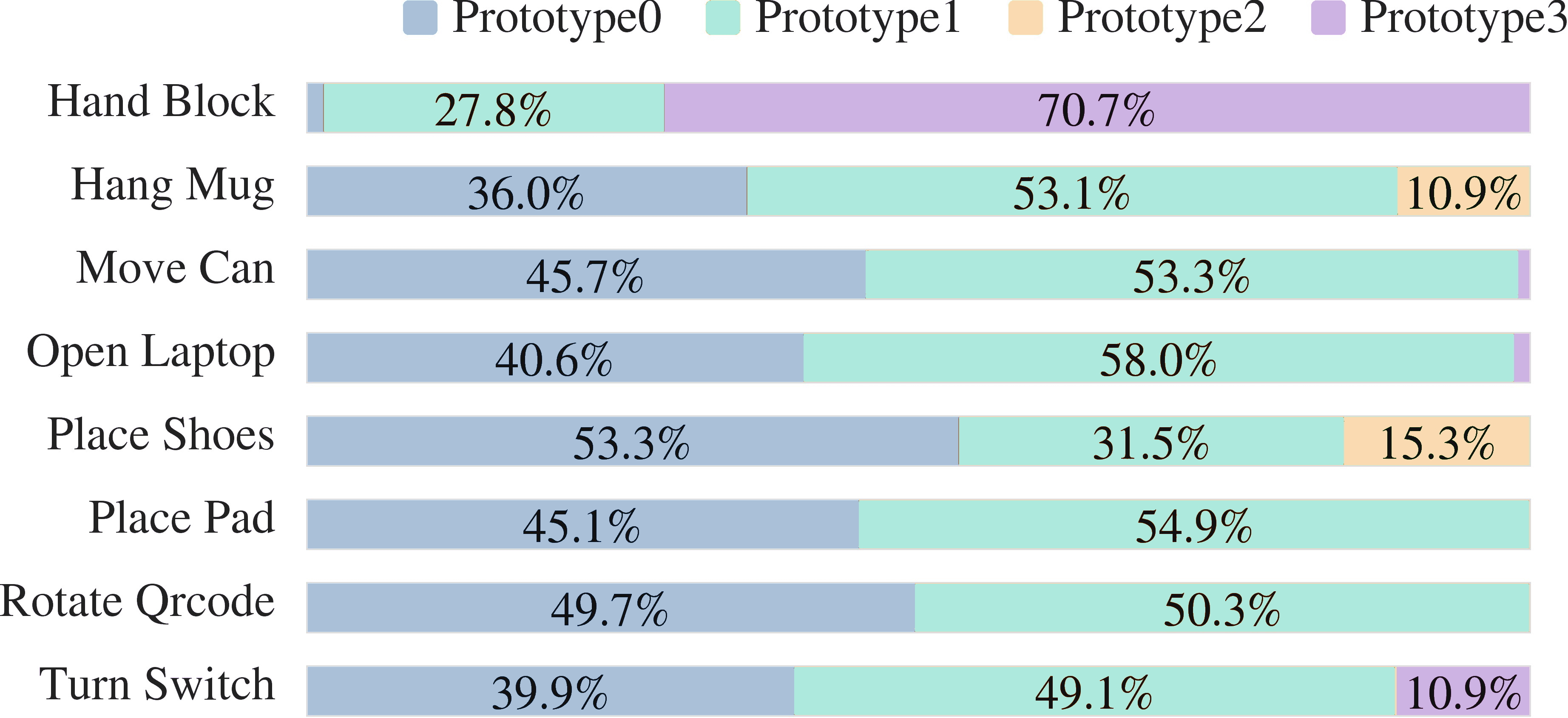}
\vspace{-0.50cm}
\caption{Illustration of kinematic archetype collapse.}
\label{Fig:kinematic_archetype}
\vspace{-0.5cm}
\end{figure}

However, the MoE-augmented VLAs hinge critically on whether the router can achieve \emph{appropriate} expert assignment when facing the fundamental gap between the action kinematics and the visual-linguistic observation within VLAs. Concretely, visually near-identical scenes may correspond to drastically different action kinematics (e.g., ``lifting the cup'' versus ``unscrewing the cap'' share nearly identical visual inputs but exhibit completely different kinematic profiles), while visually dissimilar scenes may share isomorphic kinematic patterns (e.g., ``pushing a plate'' and ``pushing a book'' differ in visual appearance but are kinematically equivalent). This fundamental semantic misalignment causes routers to suffer from physically inappropriate expert assignment if merely based on visual-linguistic observations. 

\begin{figure*}[t]
\includegraphics[width=\linewidth]{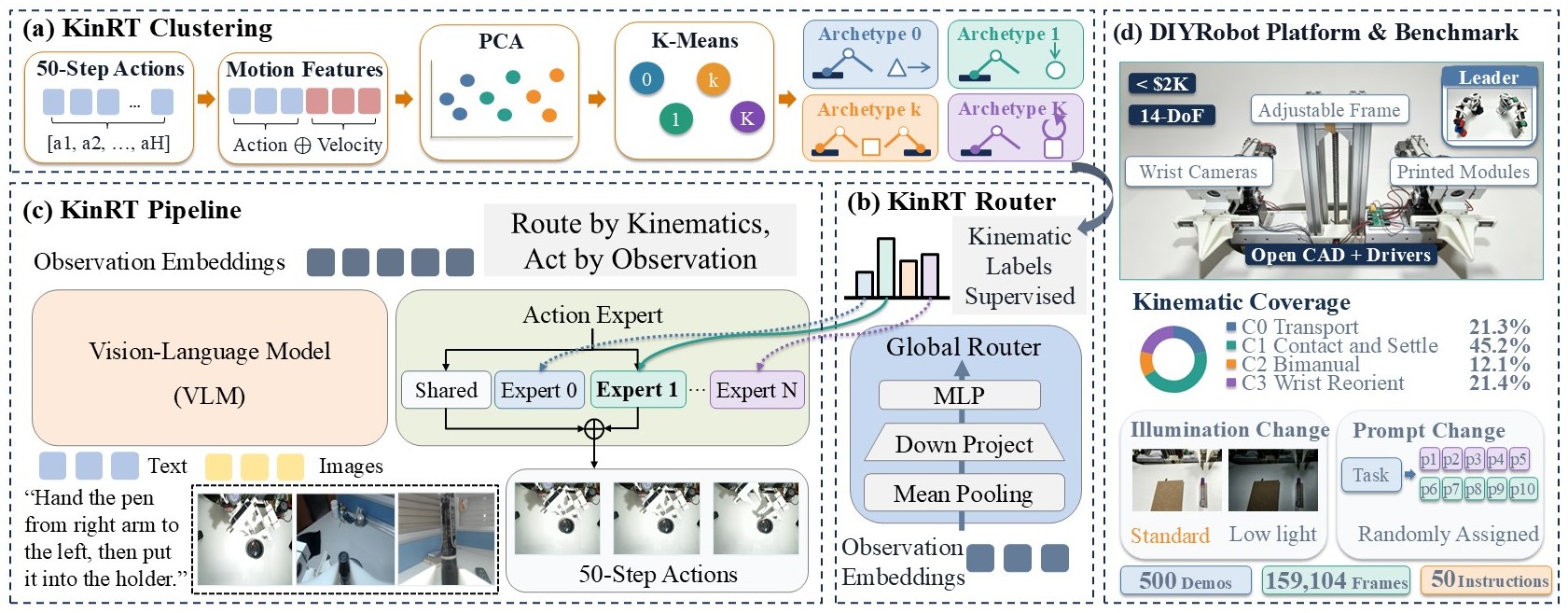}
\vspace{-0.6cm}
\caption{Overview of the proposed KinRT paradigm and our newly introduced DIYRobot platform.}
\label{Fig:KinRT_overview}
\vspace{-0.36cm}
\end{figure*}

The root cause of this expert routing failure can be traced to the implicit routing. Specifically, in conventional MoEs, the routing is determined entirely by backpropagation without any explicit consideration \cite{ref:scalingvit, mustafa2022multimodal}. Even in natural language processing, implicit routing has been observed to produce uneven expert utilization and semantically ambiguous specialization~\cite{zoph2022st}. In embodied manipulation, where kinematic heterogeneity is far more pronounced, the problem of implicit routing is exacerbated. This is because the true criterion for ``which expert should be assigned'' is neither linguistic similarity nor visual resemblance, but \emph{kinematic semantic isomorphism}. Yet, action kinematic information is fundamentally unavailable at inference time, constituting an intrinsic information asymmetry between training and deployment.

To bridge this gap, we first uncover the phenomenon of kinematic prototype collapse (as illustrated in Figure~\ref{Fig:kinematic_archetype}) via experiments, which indicates that most robot manipulation tasks eventually converge to multiple kinematic archetypes. Inspired by this observation, we propose Kinematics-supervised explicit routing (KinRT), a new paradigm that shifts from implicit, observation-driven expert routing to explicit, kinematics-supervised expert dispatching. Specifically, we leverage the discovered kinematic archetypes that serve as the ground truth to explicitly train the router through the following three integral stages: (i) we perform kinematic clustering on action trajectories to obtain multiple semantically coherent kinematic expert groups; (ii) we use the resulting cluster IDs as supervisory labels to train the router, enabling it to predict kinematic archetypes solely from visual-language observations; and (iii) at inference time, the router automatically dispatches experts based exclusively on the observed visual-language inputs, requiring no reliance on action kinematic priors. This ``train with action kinematic clustering, infer with vision-language observations'' asymmetric bridging mechanism essentially distills the structural understanding of task kinematics from the privileged action space into the kinematics-impoverished observation space. To evaluate KinRT's cross-platform generalizable capabilities, we specifically construct a Do-It-Yourself robot (DIYRobot) platform from scratch using 3D-print technology, and derive a benchmark (also named DIYRobot) based on DIYRobot platform. This DIYRobot benchmark consists of five manipulation tasks and each of them contains 100 demonstrations. The proposed KinRT, the DIYRobot platform, and the benchmark are illustrated in Figure \ref{Fig:KinRT_overview}.

In summary, the main contributions of this paper are:
\begin{itemize}
    \item We discover the phenomenon of kinematic prototype collapse, and further propose KinRT that leverages kinematic clustering IDs to supervise the router's training, and transfers the dispatching capabilities to inference where the router operates solely on visual-language observations, effectively bridging the information asymmetry gap.
    \item We deliberately build an economical and practical DIYRobot platform from scratch using 3D-print technology, which is especially suitable for small research teams in research community, and additionally collect a DIYRobot benchmark. All the required materials of DIYRobot platform and DIYRobot benchmark will be released publicly. 
    \item Extensive experiments demonstrate KinRT's substantial strengths over dense and MoE-featured VLAs by more than 23.26\% on RoboTwin benchmark and 20.27\% on DIYRobot platform, thereby validating its effectiveness.
\end{itemize}

\section{Related Work}
\label{sec:related_work}

\subsection{VLA Models for Robot Manipulation}

VLA paradigm aims to unify perception, reasoning, and action generation within a single architecture. Early efforts such as RT-1~\cite{brohan2022rt1} and RT-2~\cite{brohan2023rt2} demonstrated that policies trained on large-scale robotic data can generalize across diverse manipulation tasks. Multi-modal approaches such as VIMA~\cite{jiang2023vima} and multi-task transformers like Perceiver-Actor~\cite{shridhar2023perceiver} have additionally explored unified architectures for robot manipulation. More recent works like OpenVLA~\cite{kim2024openvla}, $\pi_0$~\cite{black2024pi_0}, $\pi_{0.5}$ \cite{intelligence2025pi_}, LingBot VLA \cite{wu2026pragmatic,lingbotvla2} have further scaled this paradigm by leveraging pre-trained backbones, diverse cross-embodiment datasets, and flow-matching action heads. Despite their impressive capabilities, these models assume that they can adequately accommodate the heterogeneous kinematics across diverse manipulation tasks. However, this assumption becomes increasingly unreliable as task diversity scales. Our work addresses this limitation by introducing kinematics-guided MoE expert routing.

\subsection{MoE and Its Routing}

MoE is a pivotal method for scaling model performance via harnessing experts' specialization across diverse data \cite{ref:moesurvey,ref:scalingvit,fan2022m3vit,shen2025expertise,du2025himoe}. The modern sparsely-gated MoE was popularized by \cite{shazeer2017outrageously}. Subsequent efforts \cite{fedus2022switch,ref:gshard} stabilized the routing mechanism. GLaM \cite{ref:glam} showed that MoE models can match or even exceed dense counterparts at a small training and inference cost. Recent open MoE systems such as Mixtral \cite{ref:mixtral} and DeepSeekMoE \cite{ref:deepseekmoe} have further refined expert granularity for practical deployment. A primary difficulty in MoE training is load collapse, where a small subset of experts dominates routing \cite{ref:baselayers,ref:expertchoice}. To solve it, BASE Layers \cite{ref:baselayers} cast token-to-expert assignment as a balanced linear assignment problem, while Expert Choice routing \cite{ref:expertchoice} inverts the selection so that experts choose tokens to guarantee balanced loads. However, routers in these works are learned implicitly via gradients, with no explicit signal to guide expert routing. Recent works~\cite{zuo2022taming} have observed that such implicitly learned routers often produce ambiguous expert assignments. Our work improves this implicit routing by incorporating explicit kinematic supervision to the router.

\subsection{Privileged Kinematic Structure in Robot Learning}

The asymmetric bridging mechanism in the proposed KinRT is conceptually rooted in the learning using privileged information (LUPI) \cite{vapnik2009new}. The core idea is that information available exclusively during training can improve generalization, even when this information is unavailable at inference time. This paradigm has been adopted in robotics where a privileged teacher with full state supervises a sensor-only student policy \cite{lee2020learning}. While prior LUPI applications in robotics typically use privileged information to supervise the policy itself \cite{liang2024rapid}, our work treats action trajectory kinematics as privileged information that supervises the router rather than the policy. This shift from action prediction to expert assignment injects task kinematics into the routing mechanism while preserving an observation-only inference pipeline.

\section{Methodology}
\label{sec:method}

\subsection{KinRT's Architecture and Design Philosophy}
\label{sec:method:moe}

Let an observation be $o=(\mathcal{I},\ell)$, comprising multi-view images $\mathcal{I}$ and a language instruction $\ell$. Our policy adopts a Mixture-of-Transformers architecture with $L = 18$ Transformer blocks. In each block, a prefix stream encodes $\mathcal{I}$ into vision tokens by a frozen SigLIP and $\ell$ into language tokens through a frozen PaliGemma-2B, while a suffix stream generates action sequence through integrating Gemma-300M and the introduced MoE extension. The two streams interact through attention over the concatenated prefix-suffix sequence while maintaining stream-specific attention projections and feed-forward network (FFN) parameters. Prefix tokens attend bidirectionally within the visual-language context. Action tokens attend to the full prefix and to the action chunk as a whole, supporting parallel prediction of the continuous action sequence.

Each block within action head replaces its FFN with a parallel MoE composition of a shared and a routed branch,
\begin{equation}
\mathrm{FFN}_{\text{MoE}}(x) \!=\! \tfrac{1}{2}\,\mathrm{FFN}_{\text{shd}}(x) \!+\! \tfrac{1}{2}\sum\nolimits_{k=1}^{K}\tilde{w}_k\,\mathrm{FFN}_{e_k}(x),
\end{equation}
where $\{e_k\}_{k=1}^{K}$ are the Top-$K$ selected experts and $\{\tilde{w}_k\}_{k=1}^{K}$ are their renormalized routing weights ($\sum_k\tilde{w}_k=1$). The shared branch is the pretrained FFN, while the $N$ expert FFNs (each matching the backbone width, $d_{\text{model}}=1024,\ d_{\text{mlp}}=4096$) are trained accordingly. This residual design keeps the shared branch permanently active, and supplies a stable starting point before the MoE experts converge. This lets each expert learn only an incremental specialization rather than relearning generic capability from scratch. The fixed weight $(1/2,1/2)$ is a uniform prior and avoids unbalanced initialization in magnitude. 

We build KinRT on the above model architecture and augment its action generation via a kinematics-supervised routing MoE. The primary problem we address is an information asymmetry: the kinematic structure of the action that genuinely determines which expert should be activated is available during training but absent at inference, where only visual-linguistic observations remain. Therefore, KinRT fills this gap through a three-stage asymmetric bridge: (i) \textbf{Kinematic Archetype Clustering:} we discover a small set of kinematic prototypes by clustering action trajectories offline. (ii) \textbf{Kinematics-Supervised Global Router:} we convert those clustering IDs into supervisory labels and train a global router to select the activated experts. (iii) \textbf{Action Generation by Observation at Deployment:} at deployment, the router dispatches experts purely based on the visual-language observation, with no dependence on action priors.

\subsection{Kinematic Archetype Clustering}
\label{sec:method:prototype}

We obtain kinematic prototypes directly from action kinematic clustering. To clearly cluster kinematic archetypes for each action position, we combine the future $H$-steps' action into an action chunk $a_0\in\mathbb{R}^{H\times D}$ where $D$ represents the space dimension of each action structured as $[\text{Left Arm}\times 6\,|\,\text{Left Gripper}\times 1\,|\,\text{Right Arm}\times 6\,|\,\text{Right Gripper}\times 1]$. In our design, by setting the horizon $H=50$, the built action chunk is $a_0\in\mathbb{R}^{50\times 14}$, which jointly captures where the arms go and how they get there. Subsequently, this action chunk is flattened into a tensor with $H\times D = 700$ dimensions to serve as position features, and the temporal difference between adjacent actions is flattened into a tensor with $(H-1)\times D = 686$ dimensions to serve as velocity features. Their concatenation yields a $1386$-dimensional descriptor
\begin{equation}
\phi_i = \big[\,\mathrm{vec}(a^{(i)}_{0:H})\;\big\|\;\mathrm{vec}(a^{(i)}_{1:H}-a^{(i)}_{0:H-1})\,\big]\in\mathbb{R}^{1386},
\end{equation}
where $i$ is the action step index. We standardize features  to remove disparities across joints and grippers, reduce dimensionality to $64$ by PCA for scalability, and finally apply frame-level $K$-means clustering.  This produces a per-frame integer prototype label $y_i\in\{1,2,\cdots, K\}$, stored offline and used to train the MoE router. Crucially, this kinematic clustering discloses the kinematic prototype collapse phenomenon (as illustrated in Figure \ref{Fig:kinematic_archetype}). 

Taking RoboTwin as an example, across the heterogeneous tasks ($162{,}545$ frames), the motion patterns collapse into four kinematic archetypes. Cluster~0 ($36.3\%$) corresponds to a left-arm-dominant early-to-mid preparation phase. Cluster~1 ($46.0\%$) captures the right-arm-dominant mid-to-late execution phase. Cluster~2 ($4.4\%$) isolates large-amplitude bimanual coordination and represents the most complex regime. Cluster~3 ($13.3\%$) is a task-specific archetype dominated by the red-block handover task. We accordingly set the number of MoE experts $N=4$ to match the kinematic granularity, so that each expert can specialize in one kinematic prototype.

\subsection{Kinematics-Supervised Global Router}
\label{sec:method:router}

We observed in experiments that cosine similarity between raw visual-language embeddings reaches above $0.95$, thereby indicating that they are non-discriminative and can not serve as the input of the MoE router. In addition, the early-step action is non-discriminative as well due to noise. We therefore use the prefix context to serve as the MoE routing input. In particular, all valid output tokens of prefix PaliGemma are aggregated by masked mean pooling (ignoring padding) into a summary vector $c\in\mathbb{R}^{2048}$. 

\paragraph{Global routing.}
Routing decisions are observation-level instead of layer-specific. For example, ``this is a red-block handover'' is equally true across all layers within the model. Therefore, we route once and broadcast the decision to all blocks' MoE. The router maps the above summarized vector $c\in\mathbb{R}^{2048}$ via a single linear layer to logits $g$. During training, we inject exploration noise $\eta\!\sim\!\mathcal{N}(0,\sigma^2)$ (disabled at inference) into $g$ and apply a temperature-scaled softmax,
\begin{align}
p=\mathrm{softmax}((g+\eta)/\tau).  
\end{align}
We then select the Top-$K$ experts based on $p$ and renormalize their probabilities into $\{\tilde{w}_k\}$. The final routing assignment $\{e_1,\cdots, e_K,\tilde{w}_1,\cdots, \tilde{w}_K\}$ is shared across all $18$ layers. Beyond efficiency (a single router pass), global routing is more stable than per-layer independent routing. Notably, we can set $\tau$ and $\eta$ accordingly.

\paragraph{Supervised routing loss.}
This is the core of the asymmetric bridge. We treat each kinematic prototype label $y_b$ as the ground truth for observation $o_b$ in training and train the router with a cross-entropy objective, i.e.,
\begin{equation}
\mathcal{L}_{\text{sup}} = -\sum\nolimits_b y_b \log \hat{y_b},
\end{equation}
where $\hat{y_b}$ is the router's predicted probability. Minimizing $\mathcal{L}_{\text{sup}}$ forces the router to recover kinematic structure from observation alone, thereby distilling the privileged action kinematic space into the visual-linguistic observation space. 

In training the global router, we introduce a sample-balancing strategy with a resampling weight $\alpha$ to mitigate the class imbalance between majority and minority kinematic archetypes. This yields a significant improvement in routing performance. Let $y_i$ be the prototype label of sample $i$ and $n_k$ be the number of samples in prototype $k$. Each sample is assigned a sampling weight $w_i = n_{y_i}^{-\alpha}$, and mini-batches are drawn with replacement according to $p_i = w_i / \sum_j w_j$. This gives the marginal prototype sampling probability $P(y = k) = n_k^{1-\alpha} / \sum_c n_c^{1-\alpha}$. The coefficient $\alpha$ interpolates between empirical sampling ($\alpha = 0$) and uniform sampling ($\alpha = 1$). We use $\alpha = 0.5$ to ensure minority-prototype exposure while retaining part of the natural data distribution.

\subsection{Action Generation by Observation at Deployment}

For each clean action $a_0$, we draw noise from a Gaussian distribution $\varepsilon \sim \mathcal{N}(0, I)$ and a time step from a Beta distribution $t \sim \mathrm{Beta}(1.5, 1)$. We then construct the noisy action
\begin{equation}
x_t = t \cdot \varepsilon + (1 - t)\, a_0.
\end{equation}
The target velocity is the time-derivative of $x_t$, i.e.,
\begin{equation}
v_t = {\partial x_t}/{\partial t} = \varepsilon - a_0,
\end{equation}
which is constant and yields a simple and stable regression target. The action generator is optimized by a MSE loss
\begin{equation}
\mathcal{L}_{act} = \mathbb{E}_{t,\,\varepsilon,\,a_0}\left\| \hat{v}_\theta(x_t, t, o) - v_t \right\|^2 .
\end{equation}

At deployment, the router operates exclusively on visual-linguistic observation $o$, completing the bridge from the action space used in training to the kinematics-absent observation space used in inference. The inference of each observation is split into a one-time precomputation and the denoising loop. In the precomputation, the visual-linguistic input passes once through the prefix stream. This produces the pooled context $c$ and yields a global routing decision $\{e_1^\ast,\cdots, e_K^\ast,\tilde{w}_1^\ast,\cdots,\tilde{w}_K^\ast\}$. The subsequent denoising loop runs for $T$ steps, each step invoking the routed action experts to predict $\hat{v}_t$. By using the learned velocity field $\hat{v}_t$, each step updates the action with increment of $\Delta t=1/T$ as
\begin{equation}
x_{t-\Delta t} = x_t - \Delta t\, \hat{v}_t.
\end{equation}
Since the learned velocity $\hat{v}_t \approx v_t = \varepsilon - a_0$ points toward noise, subtracting it pushes the updates toward the clean action. Finally, after $T$ steps the result converges approximately to the clean action sequence $x_0$.

\begin{table*}[tp]
\footnotesize
\setlength{\tabcolsep}{0.8pt}
\renewcommand{\arraystretch}{0.8}
\begin{tabular*}{\linewidth}{@{\extracolsep{\fill}}l|*{8}{w{r}{0.0234\linewidth}@{\hspace{2pt}\makebox[0pt][c]{\textbar}\hspace{-0.6pt}}w{l}{0.0234\linewidth}}w{r}{0.0295\linewidth}@{\hspace{2pt}\makebox[0pt][c]{\textbar}\hspace{-0.4pt}}w{l}{0.0295\linewidth}@{\hspace{2pt}}|c@{\hspace{1pt}}c@{\hspace{1pt}}c@{\hspace{1pt}}c@{\hspace{1pt}}c@{\hspace{1pt}}c@{}}
\toprule
\multicolumn{1}{c|}{\multirow{4}{*}{Models}} & \multicolumn{18}{c|}{RoboTwin (Clean \textbar{} Random, Success \# out of 100 tests)} & \multicolumn{6}{c}{DIYRobot (Success \# out of 50 tests)} \\ \cmidrule(lr){2-19} \cmidrule(lr){20-25}
\multicolumn{1}{c|}{} & \multicolumn{2}{c}{\makecell[c]{Hand \\ Block}} & \multicolumn{2}{c}{\makecell[c]{Hang \\ Mug}} & \multicolumn{2}{c}{\makecell[c]{Move \\ Can}} & \multicolumn{2}{c}{\makecell[c]{Open \\ Laptop}} & \multicolumn{2}{c}{\makecell[c]{Place \\ Shoes}} & \multicolumn{2}{c}{\makecell[c]{Place \\ Pad}} & \multicolumn{2}{c}{\makecell[c]{Rotate \\ Qrcode}} & \multicolumn{2}{c}{\makecell[c]{Turn \\ Switch}} & \multicolumn{2}{c|}{Avg.} & \makecell[c]{Hand \\ Pen} & \makecell[c]{Pick \\ Box} & \makecell[c]{Rotate \\ Screw} & \makecell[c]{Pull \\ Bottle} & \makecell[c]{Press \\ Button} & Avg. \\
\midrule \rowcolor{highlight}
\multicolumn{25}{l}{\textbf{\textit{Fine-tuned dense foundation models}}} \\
\multicolumn{1}{l|}{OpenVLA} & 0 & 0 & 0 & 0 & 2 & 4 & 28 & 29 & 0 & 0 & 0 & 0 & 0 & 0 & 3 & 2 & 4.1 & 4.4 & 0 & 1 & 0 & 16 & 0 & 3.4 \\
\multicolumn{1}{l|}{RDT-1B} & 19 & 9 & 6 & 0 & 16 & 12 & 46 & 33 & 1 & 1 & 1 & 1 & 16 & 12 & 2 & 4 & 13.4 & 9.0 & 0 & 3 & 0 & 6 & 4 & 2.6 \\
\multicolumn{1}{l|}{$\pi_0$-Full} & 0 & 0 & 9 & 2 & 22 & 24 & 15 & 11 & 2 & 1 & 1 & 1 & 10 & 12 & 12 & 19 & 8.9 & 8.8 & 9 & 19 & 18 & 24 & 0 & 14.0 \\
\multicolumn{1}{l|}{$\pi_0$-LoRA} & 2 & 1 & 6 & 8 & 15 & 14 & 35 & 30 & 6 & 5 & 7 & 2 & 24 & 22 & 16 & 19 & 13.9 & 12.6 & 0 & 2 & 15 & 18 & 0 & 7.0 \\
\multicolumn{1}{l|}{$\pi_{0.5}$-Full} & 5 & 3 & 4 & 3 & 26 & 27 & 76 & 81 & 11 & 2 & 4 & 10 & 40 & 31 & 31 & 28 & 24.6 & 23.1 & 19 & 31 & 33 & 39 & 26 & 29.6 \\
\multicolumn{1}{l|}{$\pi_{0.5}$-LoRA} & 8 & 21 & 8 & \bestcell{11} & \bestcell{40} & 32 & 78 & 79 & 29 & 27 & 22 & 23 & 48 & 44 & 32 & 36 & 33.1 & 34.1 & 10 & 22 & 26 & 23 & 1 & 16.4 \\
\midrule \rowcolor{highlight}
\multicolumn{25}{l}{\textbf{\textit{Fine-tuned MoE foundation models}}} \\
\multicolumn{1}{l|}{Hi-MoE} & 0 & 0 & 0 & 0 & 0 & 0 & 38 & 39 & 0 & 0 & 0 & 0 & 2 & 2 & 14 & 10 & 6.8 & 6.4 & 0 & 1 & 0 & 34 & 5 & 8.0 \\
\multicolumn{1}{l|}{AdaMoE} & 7 & 4 & 14 & 8 & \bestcell{40} & \bestcell{36} & \bestcell{95} & \bestcell{86} & 5 & 7 & 8 & 9 & \bestcell{51} & 50 & 37 & 35 & 32.1 & 29.4 & 1 & 26 & 21 & 37 & 22 & 21.4 \\
\midrule \rowcolor{highlight}
\multicolumn{25}{l}{\textbf{\textit{KinRT-augmented foundation models}}} \\
\multicolumn{1}{l|}{KinRT-OpenVLA} & 1 & 0 & 2 & 0 & 14 & 7 & 36 & 42 & 0 & 0 & 0 & 0 & 2 & 0 & 6 & 3 & 7.6 & 6.5 & 0 & 2 & 0 & 22 & 0 & 4.8 \\
\multicolumn{1}{l|}{KinRT-Full($\pi_0$)} & 1 & 2 & 4 & 6 & 10 & 7 & 43 & 33 & 1 & 1 & 5 & 3 & 11 & 19 & 14 & 15 & 11.1 & 10.8 & 3 & 33 & 35 & 41 & 0 & 22.4 \\
\multicolumn{1}{l|}{KinRT-LoRA($\pi_0$)} & 12 & 6 & 5 & 6 & 9 & 6 & 56 & 57 & 1 & 5 & 1 & 1 & 12 & 16 & 30 & 18 & 15.8 & 14.4 & 0 & 14 & 18 & 25 & 1 & 11.6 \\
\multicolumn{1}{l|}{KinRT-AdaMoE} & -- & -- & -- & -- & -- & -- & -- & -- & -- & -- & -- & -- & -- & -- & -- & -- & -- & -- & 2 & \bestcell{40} & 38 & 42 & 20 & 28.4 \\
\midrule \rowcolor{highlight}
\multicolumn{25}{l}{\textbf{\textit{KinRT (Ours)}}} \\
\multicolumn{1}{l|}{KinRT-Full} & \bestcell{34} & \bestcell{22} & 12 & 2 & 38 & 28 & 84 & 83 & 28 & 24 & 18 & 10 & 40 & 34 & 34 & 28 & 36.0 & 28.9 & \bestcell{26} & \bestcell{40} & \bestcell{41} & \bestcell{43} & \bestcell{28} & \bestcell{35.6} \\
\multicolumn{1}{l|}{KinRT-LoRA} & 18 & 17 & \bestcell{19} & 10 & \bestcell{40} & 34 & 84 & 82 & \bestcell{44} & \bestcell{41} & \bestcell{34} & \bestcell{32} & 44 & \bestcell{51} & \bestcell{43} & \bestcell{43} & \bestcell{40.8} & \bestcell{38.8} & 5 & 32 & 38 & 40 & 4 & 23.8 \\ \bottomrule
\end{tabular*}
\vspace{-0.2cm}
\caption{Performance comparison of our proposed KinRT against multiple baselines on RoboTwin and DIYRobot benchmarks.}
\label{tab:main_results}
\vspace{-0.3cm}
\end{table*}

\subsection{DIYRobot Platform and Benchmark}
\label{sec:method:diyrobot}

To assess KinRT's cross-platform generalization, we purposely construct a 14-DoF DIYRobot platform from scratch using 3D-print technology. Our DIYRobot platform is economical and practical, costing less than 2,000 USD. Its relatively low construction cost is quite suitable for small academic teams and resource-constrained labs in institutes, colleges, and universities. Using this platform, we collect a corresponding real-world manipulation benchmark, also named DIYRobot. This DIYRobot benchmark consists of five manipulation tasks (including handover pen, pick box, rotate screwdriver, pull bottle, and press button) and each of them contains 100 recorded demonstrations. We will publicly release the DIYRobot platform’s driver code and 3D-printable design files, together with the DIYRobot benchmark.

\section{Experiments}
\label{sec:experiments}

\subsection{Experimental Setups and Evaluation Metrics}

\subsubsection{Datasets.}

We conduct extensive experiments on the RoboTwin benchmark and DIYRobot platform. RoboTwin comprises eight robotic manipulation tasks. For training, each task provides 50 demonstrations under the clean setting and 50 under the random setting, yielding 800 demonstrations in total. For testing, each setting of these tasks is evaluated for 100 times, yielding 1,600 tests in total. The remaining settings (e.g., random seed) on RoboTwin follow the official recommendation. DIYRobot benchmark contains five manipulation tasks and each task contains 100 demonstrations, yielding 500 demonstrations in total. All models are trained on this dataset and evaluated on DIYRobot platform. Notably, DIYRobot benchmark involves only the clean setting but deliberately collects long-tail demonstrations.

\subsubsection{Implementation.} 

All models are initialized from their official pretrained checkpoints and fine-tuned via full-parameter fine-tuning, LoRA fine-tuning, or both. In the full-parameter setting, model parameters are updated on both benchmarks. In the LoRA setting, we apply LoRA with rank of 32/64 to the vision-language backbone/action expert for all models, with LoRA alpha set as 1. All policies predict action with a horizon of 50 steps in the flow-matching. All models are trained for 10,000 optimization steps with a batch size of 32 on two NVIDIA L20 GPUs. For models with different memory requirements, gradient accumulation is used to maintain the same effective batch size. 

In particular, KinRT-embedded MoE contains four experts and adopts Top-1 routing to enable one expert for each token. We train the router using a supervised objective with a loss coefficient of 0.05. To mitigate routing-class imbalance, we employ balanced sampling with a weight of 0.5. No extra load-balancing, contrastive-routing, or dead-expert regularization losses are used.

\subsubsection{Metrics.} 
For both benchmarks, we report both the per-task success count and the average success count across all tasks. 

\subsection{Main Results and Analyses}
\label{sec:exp:main}

Table~\ref{tab:main_results} compares KinRT and representative dense VLAs as well as MoE-featured VLAs on both the simulated RoboTwin benchmark and the real-world DIYRobot benchmark. We organize the analyses around five research questions (RQs).

\vspace{-0.1cm}
\paragraph{RQ1: Does KinRT outperform SOTA dense VLAs?}
Yes, and by a substantial margin. On RoboTwin, KinRT-LoRA achieves the best overall average of $40.8/38.8$, surpassing the strongest dense baseline $\pi_{0.5}$-LoRA ($33.1/34.1$) by $+7.7/+4.7$ points (i.e., $23.26\%/13.78\%$), while KinRT-Full ($36.0/28.9$) also clearly exceeds $\pi_{0.5}$-Full ($24.6/23.1$). On DIYRobot, KinRT-Full attains an average success of $35.6$, outperforming the best dense model $\pi_{0.5}$-Full ($29.6$) by $+6.0$ points (i.e., $20.27\%$) and more than doubling $\pi_{0.5}$-LoRA ($16.4$). Notably, dense models (e.g., OpenVLA, $4.1/4.4$ on RoboTwin and $3.4$ on DIYRobot) degrade severely as task kinematic heterogeneity grows, which corroborates our motivation that compressing disparate kinematics into a shared model space induces destructive parameter competition.

\vspace{-0.1cm}
\paragraph{RQ2: Does KinRT's explicit routing outperform implicit, observation-driven routing in MoE-featured VLAs?}
The comparison with MoE-featured baselines isolates the effect of explicit routing. The best implicit routing competitor AdaMoE reaches $32.1/29.4$ on RoboTwin and $21.4$ on DIYRobot, which KinRT-LoRA and KinRT-Full outperform by $+8.7/+9.4$ and $+14.2$ points, respectively. More strikingly, Hi-MoE collapses on the majority of tasks (e.g., $0/0$ on Move Can), yielding merely $6.8/6.4$ and $8.0$ overall. This collapse empirically confirms the failure mode we identified: implicitly learned routers driven purely by gradients degenerate expert assignments when facing kinematic heterogeneity, whereas supervising the router with kinematic archetype labels yields physically meaningful expert specialization.

\vspace{-0.1cm}
\paragraph{RQ3: Is KinRT an architecture-agnostic paradigm that can be plugged into diverse backbones?}
The KinRT-augmented variants demonstrate consistent plug-and-play gains across heterogeneous foundations. Specifically, KinRT-LoRA($\pi_0$) improves $\pi_0$-LoRA from $13.9/12.6$ to $15.8/14.4$ on RoboTwin and KinRT-Full($\pi_0$) improves $\pi_0$-Full from $8.9/8.8$ to $11.1/10.8$; KinRT-OpenVLA lifts OpenVLA from $4.1/4.4$ to $7.6/6.5$; and most notably, KinRT-AdaMoE boosts AdaMoE from $21.4$ to $28.4$ ($+7.0$) on DIYRobot. These consistent improvements indicate that the performance gains stem from the kinematics-supervised routing paradigm rather than from any specific architectural choice, and that KinRT's asymmetric bridging mechanism is broadly transferable to diverse backbone architectures.

\begin{table}[tp]
\footnotesize
\setlength{\tabcolsep}{1.66pt}
\renewcommand{\arraystretch}{0.66}
\begin{tabularx}{\linewidth}{c|ccccccccc}
\toprule
\multirow{4}{*}{$\alpha$} & \multicolumn{9}{c}{KinRT-LoRA@RoboTwin (Success \# out of 100 tests)}                                                                                       \\ \cmidrule(lr){2-10}  
                       & \makecell[c]{Hand \\ Block} & \makecell[c]{Hang \\ Mug} & \makecell[c]{Move \\ Can} & \makecell[c]{Open \\ Laptop} & \makecell[c]{Place \\ Shoes} & \makecell[c]{Place \\ Pad} & \makecell[c]{Rotate \\ Qrcode} & \makecell[c]{Turn \\ Switch} & Avg. \\ \midrule
0.0                      & 16             & 9       & \bestcell{49}       & \bestcell{84}          & 24          & 22        & 37             & 34          & 34.4 \\
0.5                      & 18             & 19       & 40       & \bestcell{84}          & \bestcell{44}          & \bestcell{34}        & \bestcell{44}             & \bestcell{43}          & \bestcell{40.8} \\
1.0                      & \bestcell{34}             & \bestcell{22}       & 45       & 72          & 25          & 17        & 39             & 39          & 36.6 \\ \bottomrule
\end{tabularx}
\vspace{-0.2cm}
\caption{Ablations on the balanced sampling coefficient $\alpha$.}
\label{tab:ablation_router_sampling_alpha}
\end{table}

\begin{table}[tp]
\footnotesize
\setlength{\tabcolsep}{0.68pt}
\renewcommand{\arraystretch}{0.66}
\begin{tabularx}{\linewidth}{l|ccccccccc}
\toprule
\multirow{4}{*}{\makecell[c]{Sources}} & \multicolumn{9}{c}{KinRT-LoRA@RoboTwin (Success \# out of 100 tests)} \\ \cmidrule(lr){2-10}
                       & \makecell[c]{Hand \\ Block} & \makecell[c]{Hang \\ Mug} & \makecell[c]{Move \\ Can} & \makecell[c]{Open \\ Laptop} & \makecell[c]{Place \\ Shoes} & \makecell[c]{Place \\ Pad} & \makecell[c]{Rotate \\ Qrcode} & \makecell[c]{Turn \\ Switch} & Avg. \\ \midrule
\makecell[l]{\makecell[c]{VLM}} & 2  & 6  & 38 & 68 & 10 & 8  & 18 & 24 & 21.8 \\
\makecell[l]{Range}  & 2  & 6  & 18 & 68 & 4  & 6  & 50 & 36 & 23.8 \\
\makecell[l]{Arm}    & 2  & 2  & 22 & 74 & 8  & 8  & 58 & 40 & 26.8 \\
\makecell[l]{Task}   & \bestcell{22} & 6 & 30 & 58 & 2 & 2 & 42 & \bestcell{46} & 26.0 \\
\makecell[l]{Velocity}    & 8  & 6  & 26 & 68 & 18 & 12 & 42 & 36 & 27.0 \\
\makecell[l]{Action}    & 14 & 12 & 20 & 68 & 34 & 12 & \bestcell{60} & 44 & 33.0 \\
\makecell[l]{\makecell[c]{Action\&\\Velocity}} & 18 & \bestcell{19} & \bestcell{40} & \bestcell{84} & \bestcell{44} & \bestcell{34} & 44 & 43 & \bestcell{40.8} \\ \bottomrule
\end{tabularx}%
\vspace{-0.2cm}
\caption{Ablation on what sources are best suitable to cluster for providing the MoE router's training label.}
\label{tab:ablation_clustering_methods}
\vspace{-0.3cm}
\end{table}

\vspace{-0.1cm}
\paragraph{RQ4: How does KinRT behave across the simulation-to-reality gap, and what is the trade-off between LoRA and full fine-tuning?}
An instructive dichotomy emerges between the two benchmarks. In simulation, KinRT-LoRA dominates ($40.8/38.8$ vs.\ $36.0/28.9$ for Full), while on the real DIYRobot platform the ordering reverses: KinRT-Full achieves $35.6$ against KinRT-LoRA's $23.8$. The same pattern holds for the $\pi_{0.5}$ baselines ($33.1$ vs.\ $24.6$ in simulation; $16.4$ vs.\ $29.6$ on real platform). This indicates a benchmark-level rather than method-level phenomenon: real-robot data collected on DIYRobot deviates substantially from the pretraining distribution, so the larger adaptation capacity of full fine-tuning becomes necessary to absorb the embodiment gap, whereas in simulation the parameter-efficient LoRA regularization mitigates overfitting to the limited demonstrations. Importantly, KinRT delivers the best result under both regimes, evidencing genuine cross-platform generalizability.

\begin{figure}[tp]
\includegraphics[width=\linewidth]{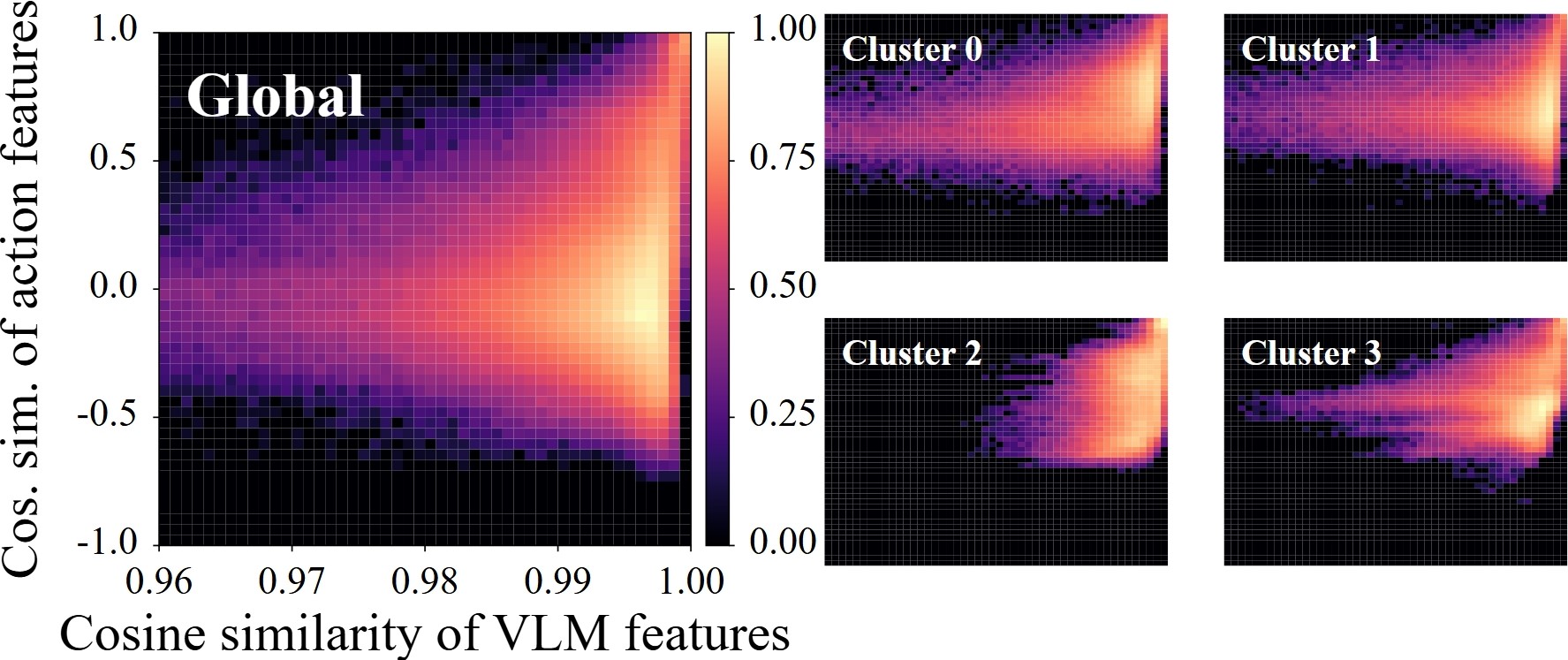}
\vspace{-0.6cm}
\caption{Demonstrations of the relationship between action-velocity space and visual-linguistic observation space.}
\label{fig:action_vel_to_observation}
\vspace{-0.38cm}
\end{figure}

\begin{figure*}[t]
\includegraphics[width=\linewidth]{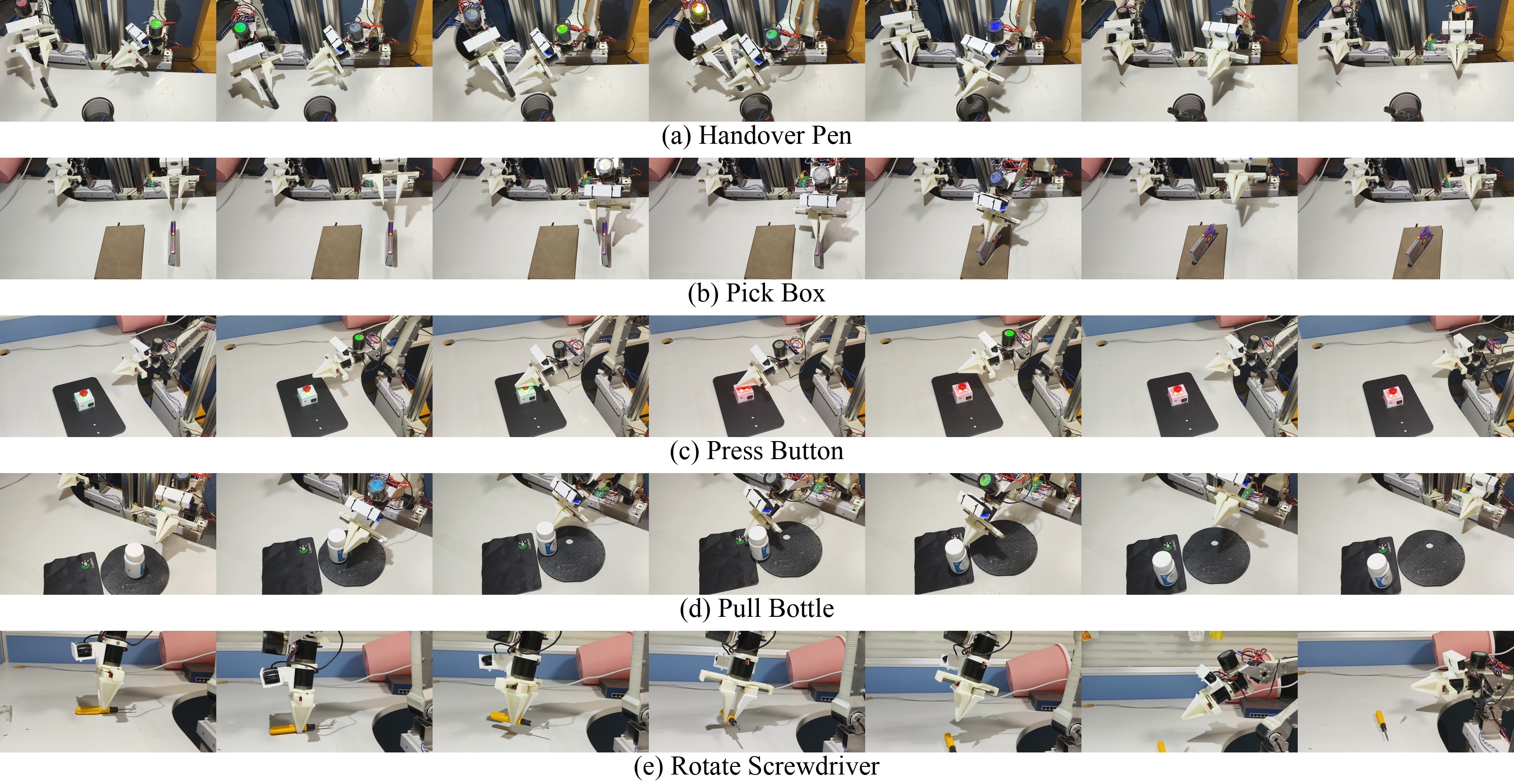}
\vspace{-0.6cm}
\caption{Demonstrations of the five manipulation tasks performed on our DIYRobot platform, where the left-to-right sequence indicates the temporal progression of each operation.}
\label{fig:diyrobot_case_studies}
\vspace{-0.38cm}
\end{figure*}

\vspace{-0.1cm}
\paragraph{RQ5: On which kinematic regimes does KinRT gain the most?}
On bimanual-coordination tasks such as Handover Block, KinRT-Full reaches $34/22$ whereas all dense and MoE baselines remain below $19$ in the Clean setting; on the real-world Handover Pen task, KinRT-Full achieves $26$ successes while most baselines except $\pi_{0.5}$-Full ($19$) remain low. On the other hand, contact-precise tasks such as Press Button, where $\pi_0$ variants and $\pi_{0.5}$-LoRA fail almost entirely ($0$--$1$), KinRT-Full attains the best result of $28$. These tasks correspond to the rare, large-amplitude bimanual and fine-positioning kinematic archetypes uncovered by our clustering. Under implicit routing MoEs or dense models, such minority regimes are overwhelmed by the dominant motion patterns, whereas KinRT's explicit supervision allocates dedicated expert capacity to them. Meanwhile, KinRT remains competitive on prototype-shared tasks (e.g., Open Laptop: $84/83$), showing that expert specialization is achieved without sacrificing performance on common kinematic regimes.

\paragraph{Summary.}
Across both benchmarks, KinRT (i) establishes new SOTA averages on both benchmarks, (ii) consistently improves diverse backbones as a plug-in, and (iii) yields the largest gains on kinematically rare and demanding tasks, jointly validating that kinematic isomorphism rather than observation similarity is the correct criterion for expert routing.

\subsection{Ablation Studies}

Table~\ref{tab:ablation_router_sampling_alpha} ablates the balanced sampling coefficient $\alpha$. The intermediate setting $\alpha=0.5$ achieves the best average success (40.8), clearly outperforming both extremes (34.4 for $\alpha=0$ and 36.6 for $\alpha=1$). This reveals a fundamental trade-off. On the one hand, without balancing ($\alpha=0$), experts corresponding to rare kinematic prototypes are under-trained due to the skewed cluster distribution, degrading tasks that rely on minority archetypes. On the other hand, with full sampling ($\alpha=1$), the empirical data distribution is over-distorted, over-fitting tail prototypes while sacrificing performance on dominant motion patterns (e.g., Open Laptop drops from 84 to 72). The compromise at $\alpha=0.5$ mitigates expert under-training while largely preserving the natural data statistics, and we therefore adopt it as the default.

Table~\ref{tab:ablation_clustering_methods} ablates the source used to cluster demonstrations into archetype labels for router supervision, and the results reveal that kinematics-derived signals dominate semantic or coarse statistical ones. Clustering on VLM embeddings (i.e., visual-linguistic similarity) performs worst ($21.8$), empirically confirming our core hypothesis that semantic proximity does not imply kinematic isomorphism. Coarse kinematic abstractions such as action range ($23.8$) and arm presence ($26.8$) improve over VLM but remain limited, as they capture only laterality of motion while discarding its temporal structure. Likewise, manual task-identity labels ($26.0$) fail to merge kinematically equivalent tasks or separate heterogeneous ones, despite occasional per-task wins (e.g., Handover Block, $22$). Among fine-grained signals, action trajectories ($33.0$) outperform velocity profiles ($27.0$), suggesting that absolute action configurations carry more discriminative archetype information than velocity. Crucially, combining action and velocity ($40.8$) yields a substantial $+7.8$ gain over the best single source and achieves the best result, indicating that action and velocity encode complementary spatial configuration and motion tempo and their joint clustering produces the most physically coherent expert partition. Overall, this ablation substantiates that complementary motion descriptors constitute the most suitable clustering source.

Figure~\ref{fig:action_vel_to_observation} visualizes the joint density of pairwise cosine similarities between pooled VLM features and PCA-projected action-velocity features, and it provides a mechanistic explanation for the ablation results in Table~\ref{tab:ablation_clustering_methods}. Two observations stand out. First, the VLM similarity axis is severely collapsed: virtually all demonstration pairs fall within the narrow band $[0.96, 1.00]$, meaning that the VLM representation is highly collapsed, with most demonstration pairs exhibiting near-identical cosine similarity. Second, and more critically, at any fixed VLM similarity, the kinematic similarity spans nearly the entire range $[-0.75, 1.0]$, with the global density mass centered around zero. This pattern explains why the VLM-clustered router performs worst ($21.8$ in Table~\ref{tab:ablation_clustering_methods}). In contrast, each per-cluster panel exhibits a markedly compacted kinematic similarity distribution. That indicates that our clustering carves the demonstration space into groups that are internally coherent in motion, confirming that the kinematic archetypes recovered by KinRT constitute the appropriate supervisory signal for expert routing.

\subsection{Case Studies}
Figure~\ref{fig:diyrobot_case_studies} presents qualitative case studies of KinRT executing the five manipulation tasks on our real-world DIYRobot platform. In \textbf{(a) Handover Pen}, one arm grasps the pen from the tabletop and smoothly transfers it to the other arm, requiring precise bimanual coordination and accurate inter-gripper alignment. In \textbf{(b) Pick Box}, the robot approaches the upright box, secures a stable grasp, and places it flat onto the target book, involving a grasp-and-reorient motion. In \textbf{(c) Press Button}, the gripper descends onto the button box and presses the button, which demands fine-grained vertical positioning. In \textbf{(d) Pull Bottle}, the robot grasps the bottle on the pad and pulls it laterally to the designated region, testing horizontal dragging with sustained contact. In \textbf{(e) Rotate Screwdriver}, the robot picks up the screwdriver and performs a wrist-dominated rotation, representing the most dexterous motion archetype among the five tasks. As shown in the frame sequences, KinRT produces smooth, temporally coherent trajectories and successfully completes all tasks despite the diverse kinematic patterns they entail, ranging from bimanual handover and coarse pick-and-place to contact-rich pulling and fine rotational manipulation. This further demonstrates the effectiveness and robustness of kinematics-guided expert routing in real-world settings.

\section{Conclusion}
\label{sec:conclusion}

We presented KinRT, a kinematics-guided MoE framework for robotic manipulation. By clustering demonstration trajectories into kinematic prototypes and using them to supervise a lightweight router, KinRT decomposes the policy into an always-active shared branch that preserves generic pretrained skills and a set of routed experts that specialize in distinct motion archetypes. Extensive experiments on the RoboTwin benchmark and our introduced real-world DIYRobot platform and benchmark demonstrate that KinRT consistently improves performance over dense and MoE-featured VLAs. Notably, KinRT achieves these gains with negligible additional inference cost. In the future, we plan to explore adaptive expert activation for kinematically heterogeneous tasks and extend kinematic routing to larger-scale VLA models.

\end{document}